\documentclass[11pt,a4paper]{article}
\usepackage[hyperref]{acl2019}
\usepackage{times}
\usepackage{latexsym}

\usepackage{url}

\usepackage{graphicx}
\usepackage{amsmath}
\usepackage{amsfonts}
\usepackage{array}
\usepackage{tabu}
\usepackage{multirow}

\aclfinalcopy 

\title{Memorizing All for Implicit Discourse Relation Recognition}

\author{Hongxiao Bai$^{1,2,3}$, Hai Zhao$^{1,2,3,}$\thanks{\;\;Corresponding author.}\;\,, Junhan Zhao$^4$ \\
  $^1$Department of Computer Science and Engineering, Shanghai Jiao Tong University \\
  $^2$Key Laboratory of Shanghai Education Commission for Intelligent Interaction \\
  and Cognitive Engineering, Shanghai Jiao Tong University, Shanghai, China \\
  $^3$MoE Key Lab of Artificial Intelligence, AI Institute, Shanghai Jiao Tong University \\
  $^4$Computer Graphics Technology, Purdue University, West Lafayette, Indiana, USA \\
  {\tt baippa@sjtu.edu.cn, zhaohai@cs.sjtu.edu.cn, zhao835@purdue.edu} 
  \\}

\date{}

\begin{document}
\maketitle
\begin{abstract}
  Implicit discourse relation recognition is a challenging task due to the absence of the necessary informative clue from explicit connectives.
  The prediction of relations requires a deep understanding of the semantic meanings of sentence pairs.
  As implicit discourse relation recognizer has to carefully tackle the semantic similarity of the given sentence pairs and the severe data sparsity issue exists in the meantime, it is supposed to be beneficial from mastering the entire training data.
  Thus in this paper, we propose a novel memory mechanism to tackle the challenges for further performance improvement.
  The memory mechanism is adequately memorizing information by pairing representations and discourse relations of all training instances, which right fills the slot of the data-hungry issue in the current implicit discourse relation recognizer.
  Our experiments show that our full model with memorizing the entire training set reaches new state-of-the-art against strong baselines, which especially for the first time exceeds the milestone of 60\% accuracy in the 4-way task.
\end{abstract}

\section{Introduction}
Implicit discourse relation recognition is one of the critical components of discourse parsing.
This task is to identify the relationship between two adjacent discourse units (sentence or clause) without explicit connectives (e.g. \emph{because}, \emph{whereas}, etc.).
This task is tough because relation recognition requires a deep understanding of the two discourse units.
Previous works have shown that this task is instrumental to many downstream tasks such as
text summarization~\cite{gerani-EtAl:2014:EMNLP2014} and question answering~\cite{jansen2014discourse}.

The most important benchmark datasets until now for this task is Penn Discourse Treebank 2.0 (PDTB 2.0)~\cite{prasad2008penn}, in which an instance is a tetrad \{\emph{Arg$_1$, Arg$_2$, implicit connective, discourse relation}\}, where the argument pair \emph{Arg$_1$} and \emph{Arg$_2$} are discourse related sentences or clauses, the implicit connectives are annotated by humans and are not known during testing.
Implicit discourse relation recognition is to disclose \emph{discourse relation} for any given \emph{Arg$_1$} and \emph{Arg$_2$} without knowing \emph{implicit connective}.
Here is an example for the instance,
\[
\begin{split}
  &\text{[\emph{Arg$_1$}]: Never mind.}\\
  &\text{[\emph{Arg$_2$}]: You already know the answer.}\\
  &\text{[\emph{Implicit connective}]: Because}\\
  &\text{[\emph{Discourse relation}]: Cause}\\
\end{split}
\]

Numerous works have been done for this task.
\citet{lin-kan-ng:2009:EMNLP} and \citet{pitler-louis-nenkova:2009:ACLIJCNLP} first practiced conventional methods with artificial linguistic features.
Since 2015, neural networks dominate the mainstreams by introducing convolutional neural network~(CNN)~\cite{zhang-EtAl:2015:EMNLP4, qin2016conll},
recurrent neural network~(RNN)~\cite{ji-haffari-eisenstein:2016:N16-1, ronnqvist-schenk-chiarcos:2017:Short},
attention mechanism~\cite{liu-li:2016:EMNLP2016, ronnqvist-schenk-chiarcos:2017:Short},
and other network methods~\cite{qin-zhang-zhao:2016:EMNLP2016, schenk-etal-2016-really, lan-EtAl:2017:EMNLP20172, dai2018naacl, guo2018coling}.

For neural network methods, the parameters learned from the training data capture the semantic features. 
However, due to the data sparsity issue, these captured features may not well semantically link arguments and their relations, which thus will heavily affect the performance.

So in this work, we propose a novel memory component storing all the semantic representations of training instances and their corresponding relations.
During testing, the model could consult the memory component and find out the similar semantic patterns and utilize the memorized relations.
The hypothesis is that if the similar instances in the training set can be retrieved, the relations of these instances must be helpful.
The training instances are stored through their encoded representations.
These memorized instances can also be considered as a sort of knowledge source, which reflects the links between semantic representations and discourse relations.
The adopted memory component can be theoretically applied to any existing suitable models.

To implement and evaluate the memory component, we integrate the memory component into the state-of-the-art model from \citet{baiCOLING} and let the augmented model be evaluated on the benchmark PDTB 2.0, which shows that the appended memory mechanism can further promote the performance over strong baseline.

This paper is organized as follows.
Section 2 reviews the related works.
Section 3 introduces the baseline model and our proposed memory component.
Section 4 demonstrates the experiments and analyses.
Section 5 states the conclusion.

\section{Related Works}

\subsection{Implicit Discourse Relation Recognition}
Since the PDTB 2.0 corpus was released, a surge of works focusing on implicit discourse relation recognition have been proposed.
And after two shared tasks~\cite{xue-etal-2015-conll,xue-etal-2016-conll} on CoNLL are held, this task attracted more researchers.
Feature-based methods~\cite{pitler-louis-nenkova:2009:ACLIJCNLP,lin-kan-ng:2009:EMNLP,zhou2010predicting} mainly focused on extracting linguistic, or semantic features from the discourse units, or the relations between unit pairs.
Then these features are distilled and sent to a classifier for relation prediction.
\citet{lin-kan-ng:2009:EMNLP} explored several common features and their combination.
\citet{lei2018linguistic} considered some semantic and cohesion features.
Recent years, most of the works focused on using neural networks to extract the features, or  to produce more suitable representations for prediction.
\citet{braud2015comparing} found that embeddings trained with neural networks are very useful.
\citet{chen_aaai} and \citet{lei_ijcai} used the relationship between words to help the classification.
\citet{zhang-EtAl:2015:EMNLP4} and \citet{qin2016conll} used CNN to encode the discourse units to representations.
\citet{qin-zhang-zhao:2016:EMNLP2016} used a gated mechanism to enhance their classifier.
\citet{ji-haffari-eisenstein:2016:N16-1} used RNN to model sentences and used graphical models to do inference.
\citet{liu-li:2016:EMNLP2016}, \citet{ronnqvist-schenk-chiarcos:2017:Short}, and \citet{guo2018coling} deployed attention mechanism for better semantic extraction.
\citet{rutherford-etal-2017-systematic} compared several network architectures.
\citet{liu_aaai}, \citet{lan-EtAl:2017:EMNLP20172}, \citet{kishimoto2018coling}, and \citet{xu2018emnlp} tried to use extra information to help the training procedure.
\citet{dai2018naacl} put discourse units to their context and made prediction in series.

Our work is orthogonal and complementary to them.
Our motivation of introducing this is to lighten the existing data-hungry bottleneck of discourse relation recognition problem which only counts on an extremely small dataset.
As the implicit task is more semantically difficult than the explicit one, the state-of-the-art performance of the former only reaches around 50\% until the very recent days, while the latter may reach 80\% or higher.
There are a few existing works trying to alleviate the data sparsity issue.
\citet{rutherford2015improving} directly relabeled explicit instances into implicit ones by manually removing the explicit connectives.
\citet{qin-EtAl:2017:Long} used a generative adversarial training method to force the implicit module to learn from the explicit module.
\citet{xu2018emnlp} utilized active learning to lead more data into training.
In this paper, we straightforwardly apply a memory component to store all possible training instances for this challenging task, which is never explored before.

\subsection{Memory Network}
\citet{memnet_weston} first proposed memory networks to store relevant information. The memory networks can reduce the long-term forgetting issues or can be used for a knowledge base.
\citet{endmemnet_sukhbaatar} optimized the memory network and trained it end-to-end which eases the training significantly.
\citet{miller-EtAl:2016:EMNLP2016} extended this mechanism to a key-value memory for machine reading comprehension.
However, our proposed memory component is not the same.
Their memory networks are to tackle the long-term dependency issue, and the memory is used for each instance temporarily, while our memory component is used for the whole system to store the training set and is fixed after training.

\section{Method}

\begin{figure*}[ht]
  \centering
  \includegraphics[width=1\textwidth]{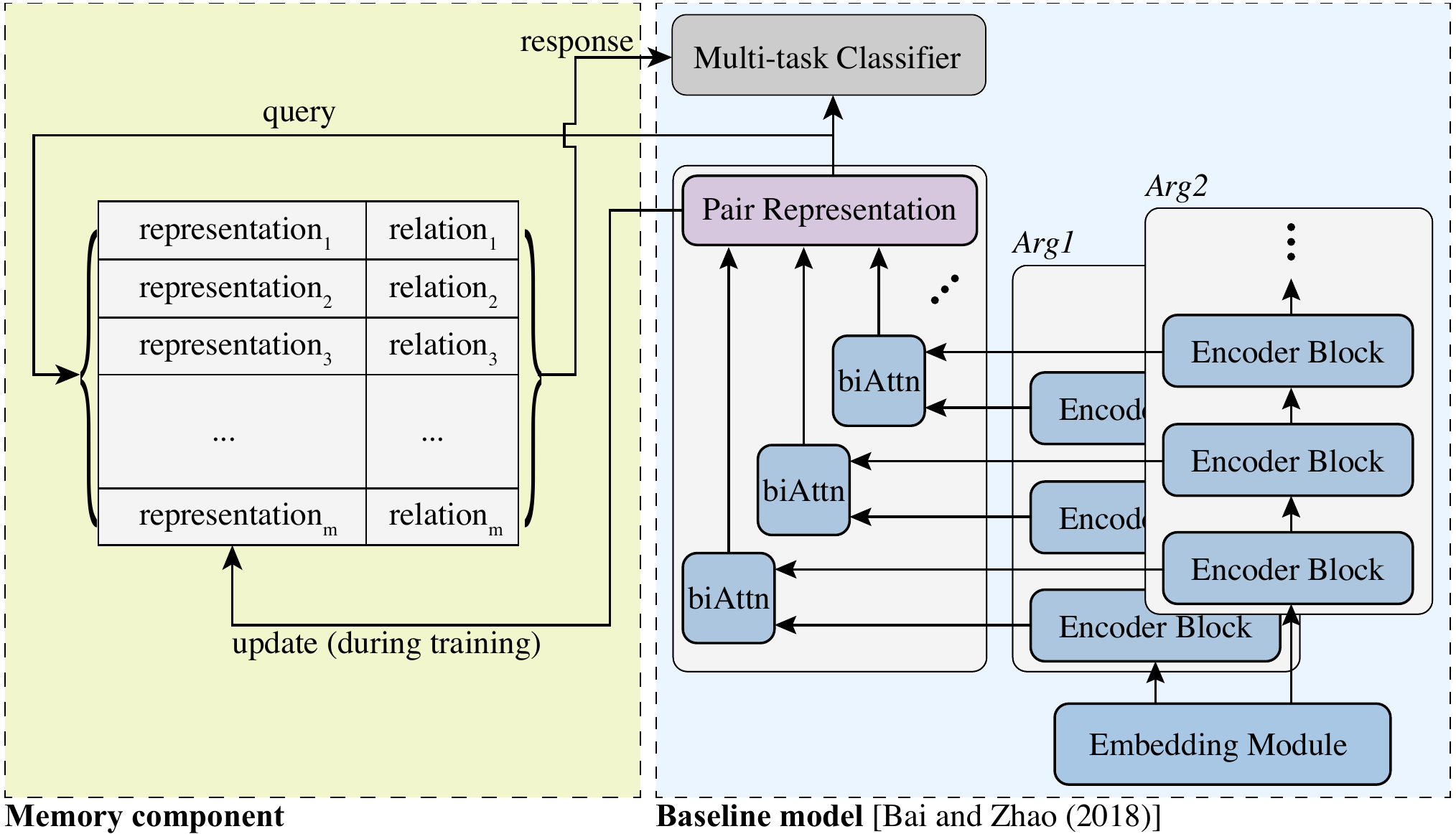}
  \caption{Model overview.}
  \label{fig:model}
\end{figure*}

The relation recognition model usually encodes the arguments first and classifies the encoded representations.
Our overall model architecture is given in Figure~\ref{fig:model}, in which a baseline module~\cite{baiCOLING} and a memory component are contained.
The baseline model is the current state-of-the-art model for this task and can provide better representations for higher performance.
Besides, we store the pair representations to the memory component to augment the whole system.

\subsection{Baseline Model}\label{sec:base}
In this section, we briefly introduce the baseline model\footnote{Please refer to the original paper for more model details.}, which reaches the current state-of-the-art results for both 11-way and 4-way implicit tasks, and consists of the following four parts.

We adopt the latest deep enhanced representation model in \citet{baiCOLING} as our baseline, which is the current state-of-the-art model which first exceeds the milestone of 48\% accuracy and 50\% $F_1$ in 11-way and 4-way implicit tasks, respectively.
Implementing this strong baseline, we store the representations of discourse unit pairs to the memory.
It provides better representations and leads to a higher performance gain.

The right part of Figure~\ref{fig:model} shows our baseline model.
It demonstrates the four parts.
The first part is the embedding module or the word-level module.
The second part is the encoding module or the sentence-level module.
The third part is the attention module or the pair-level module.
The fourth part is multi-task classifiers.

\subsubsection*{Embedding Module}
The $k$-th word of an input sentence is embedded into a vector $\mathbf{e}_k$,
which is concatenated from three parts,
$
  \mathbf{e}_k = [\mathbf{e}_k^w;~ \mathbf{e}_k^s;~ \mathbf{e}_k^c] \in \mathbb{R}^{d_e}
$.
$\mathbf{e}_k^w$ is pre-trained word embedding~(\emph{word2vec})~\cite{mikolov2013distributed}.
$\mathbf{e}_k^s$ is subword-level embedding.
$\mathbf{e}_k^c$ is the ELMo embedding~\cite{Peters2018ELMo}.

Subword units are segmented from training data using byte pair encoding~(BPE)~\cite{sennrich-haddow-birch:2016:P16-12}.
For each word, the subword sequence of the word is mapped to the subword embedding sequence.
Then convolutional operations are applied to the embedding sequence followed by max pooling operation.
Finally, the outputs are concatenated and fed to a highway network~\cite{highway} for subword-level embedding $\mathbf{e}_k^s$.

ELMo~(Embeddings from Language Models) is a pre-trained contextualized word embedding.
The outputs of this pre-trained ELMo encoder are two 1024-dimension vectors for each word.
Given this output, a self-adjusted weighted average is calculated.
Following these processes, the vector is fed to a feed forward network to reduce its dimension.

\subsubsection*{Encoding Module}
The encoding module encodes each argument separately and is composed of stacked encoder blocks.
The output of each layer is delivered to the next layer and the attention module.
\citet{baiCOLING} considers two types of encoder blocks, and we only use the convolutional type here as both types of blocks give similar performance.

Assuming the input for the encoder block is
$\mathbf{x}_k \in \mathbb{R}^{d_e} ~ (k=1, \cdots, N)$,
the input is sent to a convolutional layer and mapped to output
$[\mathbf{A}_k; \mathbf{B}_k] \in \mathbb{R}^{2d_e}$.
After the convolutional operation, a gated linear units~(GLU)~\cite{dauphin2016language} is applied, i.e.,
\[
  \mathbf{z}_k = \mathbf{A}_k \otimes \sigma(\mathbf{B}_k) \in \mathbb{R}^{d_e}
\]
There is a residual connection in this block, which adds the output and the input of the block.
Therefore, $\mathbf{z}_k + \mathbf{x}_k$ is the final output of the block corresponding to the input $\mathbf{x}_k$.
The output is delivered to the next layer and the attention module.

\subsubsection*{Attention Module}
The outputs of each layer are sent to the attention module.
Supposing the encoder block layer number is $L$, and the outputs of $l$-th block layer for \emph{Arg$_1$} and \emph{Arg$_2$} are $\mathbf{u}_1^l, \mathbf{u}_2^l \in \mathbb{R}^{N \times d_e}$, $N$ is the sentence length.
They are addressed by a bi-attention module, where the attention matrix is
\[
  \mathbf{M}_l = (\mathop{FFN}(\mathbf{u}_1^l)) {\mathbf{u}_2^l}^T
  \in \mathbb{R}^{N \times N}
\]
$\mathop{FFN}$ is a feed forward network applied to the last dimension corresponding to the word.
Then
\[
\begin{split}
  \mathbf{o}_2^l &= \mathop{softmax}(\mathbf{M}_l) {\mathbf{u}_2^l} \in \mathbb{R}^{N \times d_e}\\
  \mathbf{o}_1^l &= \mathop{softmax}(\mathbf{M}_l^T) {\mathbf{u}_1^l} \in \mathbb{R}^{N \times d_e}
\end{split}\]
the \emph{softmax} is applied to each row of the matrix.
We apply 2-max pooling on each of them and concatenate them as output
\[
  \mathbf{r}_l = [\mathop{top2}(\mathbf{y}_1^l);~ \mathop{top2}(\mathbf{o}_2^l)]
  \in \mathbb{R}^{4d_e}
\]
The final pair representation is (we let $d_r = 4{d_e}L$)
\[
  \mathbf{r} = [\mathbf{r}_1; \mathbf{r}_2; \cdots; \mathbf{r}_L] \in \mathbb{R}^{d_r}
\]
This representation is applied as the input of both the classifiers and the memory component.

\subsubsection*{Classifiers}
In this model, two classifiers are used.
One is for the relation prediction, and the other one is for the connective prediction.
\citet{qin-EtAl:2017:Long} and \citet{baiCOLING} demonstrated that connective aware information is essentially useful for the training.
The two classifiers predict the relations and the connectives simultaneously.

The classifiers are multilayer perceptrons~(MLPs) with a \emph{softmax} layer.
The connective classifier helps the model produce better representations, and only works during training. The output of the connective classifier is
\[
  \begin{split}
    \mathbf{o}_c = \mathop{softmax}[\mathop{MLP_c}(\mathbf{r})] \in \mathbb{R}^{n_c},
  \end{split}
\]
and the output of the relation classifier is
\begin{equation}
    \mathbf{o}_r = \mathop{softmax}[\mathop{MLP_r}(\mathbf{r})] \in \mathbb{R}^{n_r},
    \label{eq:rc}
\end{equation}
where $n_c$ and $n_r$ are the number of connectives and discourse relations respectively.
The loss function for both classifiers is cross entropy loss, and the total loss is the sum of the two losses
\[
  Loss = Loss_{relation} + Loss_{connective}.
\]

\subsection{Memory Component}
The left part in Figure~\ref{fig:model} shows a key-value memory component $\mathcal{S}$ to memorize the \emph{(representation, relation)} pairs of all the training instances.
The keys $\mathcal{S}^k$ are the semantic representations of discourse unit pairs in the training set, and the values $\mathcal{S}^v$ are the corresponding relations.
The keys are updated during training for better retrieval, and the memorized keys are retained for testing.

Supposing there are $m$ slots in the memory $\mathcal{S}$, which can contain $m$ training instances.
We intend to sample out $m$ training instances and index them with $1, \cdots, m$.
Therefore, the $i$-th training instance will be stored at the $i$-th memory slot $\mathcal{S}_i$.
(In our experiments, we set $m$ with the number of all training instances, namely, we memorize the entire training set.)

The key part $\mathcal{S}^k$ of the memory component is initialized randomly and is updated during training.
Simultaneously, the value part $\mathcal{S}^v$ is initialized with the one-hot encodings which represent the relations.
These encodings are fixed all the time.
Given $n_r$ relations, the one-hot representation of the $j$-th relation ($j = 1, \cdots, n_r$) is
\[
  \begin{split}
    [0,\; \cdots,\; 1,\; \cdots,\; 0] \in \mathbb{R}^{n_r},
  \end{split}
\]
that is, assign the whole vector to $0$ and the $j$-th number to 1.

\subsubsection*{Update}
The key part $\mathcal{S}^k$ of the memory is updated during training.
The baseline module can produce a pair representation $\mathbf{r}$ for classification, which is used as the semantic representation to update the key part $\mathcal{S}^k$ in the memory.

In each epoch of the training procedure, every instance in the training set will be input once.
Then one representation for each instance is produced in one epoch.
Supposing there is one pre-sampled training instance, for example, the $i$-th instance, then the corresponding output pair representation $\mathbf{r}_i$ will fill the $i$-th key slot $\mathcal{S}^k_i$ of the memory.
In each epoch, all the key slots $\mathcal{S}^k$ of the memory are updated precisely once.

After the training procedure, the memorized information in memory is fixed and can be queried during testing.
The memorized information is the correspondence between fine-tuned representations and discourse relations in the training set.
Through updating, the representations can carry semantic information about discourse arguments and are better for value retrieval.

\subsubsection*{Query}\label{sec:qr}
In this stage, the relevant memorized information is retrieved and delivered to the classifier.
For the query, each candidate in the memory is assigned to a relevance weight reflecting the semantic similarity with the query representation.
Then the useful information in the memory is retrieved by taking their weighted sum using the relevance weights for the response, which reflects the discourse relations of the most similar instances.

Supposing that the pair representation to be classified is $\mathbf{r}_q \in \mathbb{R}^{d_r}$, which is also used for the query, and the representation key in memory slot $\mathcal{S}^k_i$ is $\mathbf{r}_i \in \mathbb{R}^{d_r}$.
The query can be seen as two steps.

The first step is to assign a coefficient to each candidate in the memory component respectively.
During training, our model is kept evaluated on the whole training set.
Supposing after one of the training epochs, the number of correctly predicted training instances for the $j$-th relation~($j = 1, \cdots, n_r$) is $m_j$.
For the $i$-th training instance, if it is mispredicted, then we assign the coefficient $c_i$ with $0$.
Otherwise, if it is correctly predicted, and the relation of it is the $j$-th relation, then we assign $c_i = 1 / m_j$.
In other words, we only select out the correctly predicted instances and assign a coefficient to it to balance the result since the number of different relations is not the same.

The second step is to calculate the relevance weight corresponding to $\mathbf{r}_j$,
\begin{equation}
  w_j = f(\mathbf{r}_q, \mathbf{r}_j),
\label{eq:w}
\end{equation}
where $f$ is an attention function.
The $f$ can have different choices, such as dot product:
\begin{equation}
    f(\mathbf{r}_q, \mathbf{r}_j) = \mathbf{r}_q^T \mathbf{r}_j;
    \label{eq:dot}
\end{equation}
biaffine attention~\cite{dozat2017biaffine, jiaxun2018coling}:
\begin{equation}
    f(\mathbf{r}_q, \mathbf{r}_j) = \mathbf{r}_q^T \mathbf{U} \mathbf{r}_j + \mathbf{w}_1^T \mathbf{r}_q + \mathbf{w}_2^T \mathbf{r}_j + b,
  \label{eq:biaffine}
  \end{equation}
where $\mathbf{U} \in \mathbb{R}^{d_r \times d_r}$, $\mathbf{w}_1, \mathbf{w}_2 \in \mathbb{R}^{d_r}$ and $b \in \mathbb{R}$ are parameters;
or other attention methods.
The advanced attention mechanism such as biaffine can learn more query patterns while dot product attention returns the most similar one according to cosine distance without training.

After that, these relevance weights are normalized with \emph{softmax} and used for the response.

\subsubsection*{Response and Classification}
Supposing the value in slot $\mathcal{S}^v_i$ is $\mathbf{v}_i$, which is the one-hot vector for discourse relation, then the final response vector will be
\begin{equation}
    \mathbf{v} = \sum_{i=1}^m \mathop{softmax}(w_i) c_i \mathbf{v}_i \in \mathbb{R}^{n_r},
\label{eq_response_v}\end{equation}
where $n_r$ is the number of discourse relations introduced before.
Then this vector is sent to an MLP, and we modify the output of the relation classifier~(Eq.~\ref{eq:rc}) to
\[
  \begin{split}
    \mathbf{o}_r = \mathop{softmax}[(1-\lambda)\mathop{MLP_r}(\mathbf{r}) + \lambda\mathop{MLP_m}(\mathbf{v})],
  \end{split}
\]
where $\lambda$ is a hyperparameter and $\mathop{MLP_m}$ is the MLP for memory response vector.
This response contains the true label information and reflects the links between representations and relations.

Though $\mathbf{r}_i$~($i = 1, \cdots, m$) is used as the key in the memory component, it is also a semantic representation of the corresponding training instance.
So we can still use it as another choice for memory response vector
\begin{equation}
    \mathbf{v}\prime = \sum_{i=1}^m \mathop{softmax}(w_i) c_i \mathbf{r}_i \in \mathbb{R}^{d_r},
\label{eq_response_k}\end{equation}
then the output of the relation classifier will be
\[
  \begin{split}
    \mathbf{o}_r = \mathop{softmax}[(1-\lambda)\mathop{MLP_r}(\mathbf{r}) + \lambda\mathop{MLP_r}(\mathbf{v}\prime)],
  \end{split}
\]
here we use the MLP of relation classification for response vector.
This response is the representation from training arguments, and since we use coefficient to filter the memory, so the classifier can correctly classify them.
This response can also help the learning procedure of the classifier.

\section{Experiments\footnote{The code is available at \url{https://github.com/hxbai/IDRR_mem}}}

\subsection{Dataset Settings}
Recent works mainly use PDTB 2.0 as the benchmark dataset and we follow them.
The adopted benchmark PDTB 2.0 has three levels of relations: Level-1 \emph{Class}, Level-2 \emph{Type}, and Level-3 \emph{Subtypes}.
The first level consists of four major relation Classes and the second level contains 16 Types.
According to \citet{TACL536}, on the second level five relation types have no dev and test instances, thus they are removed, so there are 11 types in total.
We conduct evaluations on two levels: 11-way classification on level 2 and 4-way classification on level 1.

We follow the dataset settings of the previous works.
For the 11-way classification, we use two splitting methods:
the first is PDTB-Lin's splitting~\cite{lin-kan-ng:2009:EMNLP}, which uses section 2-21, 22 and 23 as training, dev and test sets respectively.
the second is PDTB-Ji's splitting~\cite{TACL536}, which uses section 2-20, 0-1, and 21-22 as training, dev and test sets respectively.
For 4-way classifications, the splitting is the same as the PDTB-Ji's splitting in 11-way classification without eliminating instances.

During training, the instances with more than one annotated relation types are considered as multiple instances.
At test time, a prediction matching one of the gold types is taken as the correct answer.
All sentences in the dataset are padded or truncated to keep the uniform 100-word length.

\subsection{Model Details}

Most of the hyperparameter settings in our experiments are the same as the baseline model of \citet{baiCOLING}.
For Lin's splitting, we change the layer number of classifiers to 2, the hidden dim of the classifiers to 2048, and the learning rate of it to 0.0012.
We set $\lambda = 0.3$ and the memory dropout to 0.2.

\subsection{Results}

\begin{table*}[ht]
  \centering
  \begin{tabu}{p{170pt}p{70pt}<{\centering}p{50pt}<{\centering}}
  \tabucline[0.65pt]{-}
  Model & PDTB-Lin & PDTB-Ji \\ \hline
  \citet{lin-kan-ng:2009:EMNLP}
  & 40.20 & - \\
  \citet{lin-kan-ng:2009:EMNLP} + Brown clusters
  & - & 40.66 \\
  \citet{TACL536}
  & - & 44.59 \\
  \citet{qin-zhang-zhao:2016:COLING}
  & 43.81 & 45.04 \\
  \citet{qin-EtAl:2017:Long}
  & 44.65 & 46.23 \\ 
  \citet{baiCOLING}
  & 45.73 & 48.22 \\
  \hline
  Ours
  & \textbf{46.08} & \textbf{49.15} \\ \tabucline[0.65pt]{-}
  \end{tabu}
  \caption{Accuracy~(\%) for 11-way classification.}
  \label{table:result}
\end{table*}

\begin{table*}[ht]
  \centering
  \begin{tabu}{p{130pt}cccccc}
  \tabucline[0.65pt]{-}
  Model & Comp. & Cont. & Exp. & Temp. & 4-way ($F_1$) & 4-way (Acc) \\ \hline
  \citet{rutherford2015improving}
  & 34.20 & 43.90 & 69.10 & 14.7 & 40.50 & 57.10 \\
  \citet{zhang-EtAl:2015:EMNLP4}
  & 33.22 & 52.04 & 69.59 & 30.54 & - & - \\
  \citet{TACL536}
  & 35.93 & 52.78 & - & 27.63 & - & - \\
  \citet{chen-EtAl:2016:P16-13}
  & 40.17 & 54.76 & - & 31.32 & - & - \\
  \citet{qin-zhang-zhao:2016:EMNLP2016}
  & 41.55 & 57.32 & 71.50 & 35.43 & - & - \\
  \citet{liu_aaai}
  & 34.65 & 46.09 & 69.88 & 31.82 & 44.98 & 57.27 \\
  \citet{liu-li:2016:EMNLP2016}
  & 39.86 & 54.48 & 70.43 & \textbf{38.84} & 46.29 & 57.57 \\
  \citet{qin-EtAl:2017:Long}
  & 40.87 & 54.56 & 72.38 & 36.20 & - & - \\
  \citet{lan-EtAl:2017:EMNLP20172}
  & 40.73 & \textbf{58.96} & 72.47 & 38.50 & 47.80 & 57.39 \\
  \citet{lei2018linguistic}
  & 43.24 & 57.82 & \textbf{72.88} & 29.10 & 47.15 & - \\
  \citet{dai2018naacl}
  & - & - & - & - & 48.82 & 58.20 \\
  \citet{baiCOLING}
  & \textbf{47.85} & 54.47 & 70.60 & 36.87 & 51.06 & - \\
  \hline
  Ours
  & 47.15 & 55.24 & 70.82 & 38.20 & \textbf{52.19} & \textbf{60.69} \\  \tabucline[0.65pt]{-}
  \end{tabu}
  \caption{$F_1$ score~(\%) comparison on binary classification.
  $F_1$ score~(\%) and accuracy~(\%) comparison on 4-way classification. (Only single models are compared.)}
  \label{table:result2}
\end{table*}

Table~\ref{table:result} is the comparison on 11-way classification and Table~\ref{table:result2} is the comparison on 4-way classification.
Our memory method yields performance gain and achieves a new state-of-the-art performance in both 11-way and 4-way classification.
For binary classification, three out of four results of our method are better than the baseline.

All the results are averaged from multiple runs.
For Lin's splitting, the result is achieved with the dot product attention~(Eq.~\ref{eq:dot}) and the response from values~(Eq.~\ref{eq_response_v}), and for Ji's splitting, the result is achieved with the biaffine attention~(Eq.~\ref{eq:biaffine}) and the response from keys~(Eq.~\ref{eq_response_k}).
For 4-way and binary classifications, the results are achieved with the biaffine attention~(Eq.~\ref{eq:biaffine}) and the response from keys~(Eq.~\ref{eq_response_k}).
The analyses about the choice of attention methods and responses are in the next subsection.

\subsection{Analysis}\label{sec:ana}

We have conducted some analyses and ablation studies on our memory component to illustrate the effectiveness of our method.
These experiments are conducted with the PTDB-Ji splitting on 11-way classification if not specified.

\subsubsection*{Time Consumption}

The first analysis is about the training time of our method.
We conduct a comparative experiment and find that even with the biaffine attention (which is more complex than the dot product attention), the consumed time of our new model is $7\%$ more than the baseline model since the training set is relatively small.
So the memory component only brings little impact on training time.

\subsubsection*{Key-part Strategy}

Besides the updating strategy mentioned before, we also examine a fixed key scheme to show the advantage of our method, this scheme can only be used with the value response.
In this scheme, we concatenate the pre-trained \emph{word2vec} embeddings and ELMo to a 2348-dimension vector for each word.
Then for each argument~(\emph{Arg$_1$} and \emph{Arg$_2$}), we average the word embeddings as its representation.
Then the two representations are concatenated to the pair representation.
This representation is memorized in the key part and fixed all the time.
The query vector uses the same scheme as the keys.
We use biaffine attention here.

For the query, we test two schemes. The first examines a fixed query representation, which is the same as the representations used in updating.
The second applies the encoded representation $\mathbf{r}_q$ which is introduced before, and we use a one layer MLP to make its dimension the same as the keys.
Then such a setting receives much lower accuracy 47.83\% and 48.32\% respectively compared to our dynamic key results 49.09\%, which indicates that the fixed representation fails to successfully extract features about discourse relations.
The concatenated word embeddings and ELMo are indeed semantic representations, whereas they still lack effective informative clues on the connection between argument pairs, which even makes the performance worse.
Contrarily, the dynamic keys and queries in our memory component can capture more salient relation features for better performance.

\subsubsection*{Coefficient}

As introduced in Section~\ref{sec:qr}, we assign each training instance in the memory component a coefficient, which is used to select out the correctly predicted training instances and balance the instance number of different classes.
For example, the numbers of instances of different classes in the training set of 4-way classification are 689, 3288, 1898, 6900 respectively, which are extremely unbalanced.
Without a balancing control, the class with much more instances will have an overwhelming impact.
The unbalancing issue also exists for 11-way classification.

Then, we try to fix the coefficient, that is, all the training instances can be queried during testing and the coefficient only works for balancing.
The result of this experiment is $48.93\%$, which is lower than dynamic coefficient~($49.15\%$), but higher than the baseline model~($48.22\%$).
It means that incorporating all the training instances in the memory component can bring useful information, but will have more noise than the memory filtered by the coefficient.

\subsubsection*{Attention and Response}

Section~\ref{sec:qr} mentions two attention strategies and two response methods.
Here we did several experiments for them in different settings.
The results are in Table~\ref{table:attres}.
Here we did not have results on combined key and value response (add the two types of responses as the final response) since its performance is similar to that only with the key response.

From the table, we can find the performance is heavily related to the dataset settings.
The performance gain on Lin's splitting is smallest and the results with different settings on it are extremely unstable.
With the key response, the performances are even drastically worse than the baseline.
Except for Lin's splitting, the biaffine attention with key response can achieve the best performance.

As introduced before, Ji's splitting for 11-way classification and the splitting on 4-way classification are the same, which has a larger test set than Lin's splitting.
Thus it is not surprising that a smaller test set makes the results on Lin's splitting insignificant and maybe hardly query from the memory component, or the queried information may be too noisy to promote the performance.

The choice of the combination method of the attention and the response needs to consider the dataset settings and the attention and the response can affect each other.

\begin{table}[ht]
  \centering
  \begin{tabu}{ccccc}
  \tabucline[0.65pt]{-}
  \multirow{2}{*}{Model} & \multicolumn{2}{c}{11-way} & \multicolumn{2}{c}{4-way} \\
   & Lin & Ji & $F_1$ & Acc. \\ \hline
  baseline & 45.73 & 48.22 & 51.06 & - \\ \tabucline[0.1pt on 2pt off 2pt]{-}
  D + K & 40.86 & 48.63 & 50.51 & 59.66 \\
  D + V & \textbf{46.08} & 48.99 & 51.69 & 60.39 \\
  B + K & 38.47 & \textbf{49.15} & \textbf{52.19} & \textbf{60.69} \\
  B + V & 45.92 & 49.09 & 51.22 & 60.15 \\
  \tabucline[0.65pt]{-}
  \end{tabu}
  \caption{Comparison for attention and response methods on different settings. D denotes dot product attention, B denotes biaffine attention, K denotes use key response, and V denotes use value response.}
  \label{table:attres}
\end{table}

\subsubsection*{Example}

Table~\ref{table:example} shows an example for the queried instances.
We find that the model pays attention to instances of relevant relations and the weight assigned to it is nearly $1$, that is, the model focuses on exactly one instance in the memory.
These retrieved instances indeed help the prediction of the test instance.


\begin{table}[ht]
  \centering
  \begin{tabu}{p{0.94\columnwidth}}
  \tabucline[0.65pt]{-}
  \textbf{Test instance} \\ \hline
  \textbf{Relation 1:} Contingency.Cause \\
  \textbf{Relation 2:} Expansion.List \\
  \textbf{Arg$_1$:} The HUD budget has dropped by more than 70\% since 1980. \\
  \textbf{Arg$_2$:} We've taken more than our fair share. \\
  \tabucline[0.65pt]{-}
  \textbf{Queried training instance top 1} \\ \hline
  \textbf{Relation:} Contingency.Cause \\
  \textbf{Arg$_1$:} At 11.1\% of gross national product, U.S. health costs already are the highest in the world. By contrast, Japan's equal 6.7\% of GNP, a nation's total output of goods and services.  \\
  \textbf{Arg$_2$:} Management and labor worry that the gap makes U.S. companies less competitive. \\
  \tabucline[0.65pt]{-}
  \end{tabu}
  \caption{A example for the queried training instances.}
  \label{table:example}
\end{table}

\subsection{Discussion}
The proposed memory component seems to work like the nearest neighbor method.
However, the whole component is dynamically adjusted rather than the static data setting for the nearest neighbor method.
The keys are updated with optimized pair representations.
If the keys are fixed, the performance will be worse according to our empirical verification.
Eq.~\ref{eq:w} is differentiable, which means the loss can back-propagate through the memory component to facilitate the baseline model, and if $f$ in Eq.~\ref{eq:w} has parameters (such as the biaffine attention), they can also be tuned.
In the meantime, it is not easy for the nearest neighbor method to design a dynamic distance function and a data update method, which are the right powerful designs in our propose memory mechanism according to our discussion in Section ~\ref{sec:ana}.

\section{Conclusion}
In this paper, we propose a novel memory component to enhance state-of-the-art implicit discourse relation recognition model.
The augmented memory component can memorize useful salient knowledge about the pair representations and the discourse relations in the training set.
This knowledge can benefit the prediction performance during the testing procedure.
Our system can dynamically adjust so that query and response can be better during training.
Our experiments show that putting the whole training set into the memory lets our model receive the most favorable results and achieves new state-of-the-art performance for the concerned challenging task.

\bibliography{ref}
\bibliographystyle{acl_natbib}

\end{document}